\documentclass[sigconf]{acmart}
\usepackage[capitalise, noabbrev]{cleveref}

\AtBeginDocument{%
  \providecommand\BibTeX{{%
    \normalfont B\kern-0.5em{\scshape i\kern-0.25em b}\kern-0.8em\TeX}}}

\setcopyright{acmcopyright}
\copyrightyear{2023}
\acmYear{2018}
\acmDOI{XXXXXXX.XXXXXXX}

\acmConference[ISWC]{ISWC'23: ACM International Symposium on Wearable Computer}{ October 08--12,
  2023}{Cancun, Mexico}
\acmBooktitle{ISWC'23: ACM International Symposium on Wearable Computers,
 October 08--12, 2023, Cancun, Mexico} 
\acmPrice{15.00}
\acmISBN{978-1-4503-XXXX-X/18/06}

\acmSubmissionID{7420}
\begin{document}

\title{Selecting the motion ground truth for loose-fitting wearables: benchmarking optical MoCap methods}%Are low-cost marker-less vision MoCap methods sufficient as ground truth with loose garments?}
%Are expensive MoCap systems suitable for body pose ground truth with loose garments?} %Are low-cost marker-less MoCap methods sufficient for ground truth in loose garment?}
%DrapedMoCapBench: Benchmarking 3D Human Pose Estimation Methods on different Fitting Clothes using 3D Simulated Data

\author{Lala Shakti Swarup Ray, Bo Zhou, Sungho Suh, Paul Lukowicz}
\affiliation{%
  \institution{German Reserch Center for Artificial Intelligence (DFKI) and University of Kaiserslautern, Germany}  
  \country{}
  \streetaddress{Trippstadterstr. 122}
  %\city{Kaiserslautern}
  \postcode{67663}
}
\email{{lala\_shakti\_swarup.ray, bo.zhou, sungho.suh, paul.lukowicz}@dfki.de}

\renewcommand{\shortauthors}{Ray and Zhou, et al.}

\begin{abstract}

To help smart wearable researchers choose the optimal ground truth methods for motion capturing (MoCap) for all types of loose garments, 
we present a benchmark, DrapeMoCapBench (DMCB), specifically designed to evaluate the performance of optical marker-based and marker-less MoCap. 
High-cost marker-based MoCap systems are well-known as precise golden standards.
However, a less well-known caveat is that they require skin-tight fitting markers on bony areas to ensure the specified precision, making them questionable for loose garments.
On the other hand, marker-less MoCap methods powered by computer vision models have matured over the years, which have meager costs as smartphone cameras would suffice.
To this end, DMCB uses large real-world recorded MoCap datasets to perform parallel 3D physics simulations with a wide range of diversities: six levels of drape from skin-tight to extremely draped garments, three levels of motions and six body type - gender combinations to benchmark state-of-the-art optical marker-based and marker-less MoCap methods to identify the best-performing method in different scenarios.
%Then, state-of-the-art optical marker-based and marker-less MoCap methods were applied to the simulation to identify the best-performing method in different scenarios.
%Our evaluation shows that neither marker-based or marker-less methods achieve a minimum position error of <10cm.
In assessing the performance of marker-based and low-cost marker-less MoCap for casual loose garments both approaches exhibit significant performance loss (>10cm), but for everyday activities involving basic and fast motions, marker-less MoCap slightly outperforms marker-based MoCap, making it a favorable and cost-effective choice for wearable studies. The code is available at \texttt{github.com/lalasray/DMCB/}.
\end{abstract}



 % \includegraphics[width=\textwidth]{sampleteaser}
 % \caption{Seattle Mariners at Spring Training, 2010.}
 % \Description{Enjoying the baseball game from the third-base
 % seats. Ichiro Suzuki preparing to bat.}
 % \label{fig:teaser}


\maketitle

\section{Introduction}
\vspace{-3pt}
\label{sec:intro}
Wearable sensing systems have gained a growing interest towards motion tracking, including IMU sensors \cite{gong2021robust, jiang2022transformer, yi2022physical}, RFIDs \cite{jin2018towards}, capacitive fabric sensors \cite{zhou2023mocapose}, computational fabrics \cite{liu2019reconstructing}, and multi-modalities \cite{Liu2020A}.
With continuous motion tracking, activity recognition in various scenarios is trivial downstream tasks \cite{behera2020deep, jansen20073d} along with the shared representation with other domains such as computer vision and large language models \cite{radford2021learning,moon2022imu2clip}.

However, the widely accepted golden standard for motion capture (MoCap) systems, such as Qualisys (Sweden), Vicon (USA), and OptiTrack (USA), relies on optical markers placed on the body \cite{jiang2022transformer, yi2022self}. These systems employ skin-tight marker placement on bony areas and rely on rigid biomechanical models to convert the surface points to inner joints \cite{skeletontracking, groen2012sensitivity}.
% However, the consensus golden standard ground truth MoCap systems, such as Qualisys (Sweden), Vicon (USA), and OptiTrack (USA), based on optical markers are specified with skin-tight marker placement and even on bony areas because they rely on rigid biomechanical models to convert the surface points to inner joints \cite{skeletontracking, groen2012sensitivity}.
Optical marker-based MoCap utilizes either active markers \cite{barca2006new,raskar2007prakash} with built-in light sources or passive markers \cite{lee2017low} with unique visual patterns or retro-reflective properties. 
These systems capture the surface marker positions using synchronized camera triangulation and infer the joint motion inside the human body using biomechanical models \cite{skeletontracking}. 
Optical MoCap systems are generally favored over inertial methods (that lack absolute positioning) due to their simplicity, accuracy, and robustness against external interference  \cite{fleron2019accuracy}. %and real-time tracking
However, when the markers are placed over a loose piece of garment, they are not able to follow the underlying body motion, which results in significant kinematic errors \cite{mcfadden2020sensitivity}. %as stated by McFadden et al.
The development of loose-fitting is crucial in wearable applications \cite{zhou2023mocapose, bello2021mocapaci, mcadams2011wearable} to improve user acceptance, comfort, accommodation of various body shapes, and mass adoption. 
Nevertheless, relying on marker-based MoCap to provide motion ground truth constraints further the development of loose garments.
% Constraints related to marker-based mocap limits further development of loose garments, 
%On one hand, this limits further development of loose garments, which is crucial for user acceptance; 
% On the other, studies with loose garments like casual apparel, where the optical markers need to be exposed on the outside of the garment, are in practice outside the specification of those systems.
Video-based marker-less MoCap deep learning algorithms map semantic information (e.g., body parts) to pose without explicit markers using deep learning \cite{sigal2021human, chatzis2020comprehensive,gamra2021review} have matured with the rapid advancement of artificial intelligence.
But there is a lack of comprehensive comparisons between marker-based and marker-less MoCap systems, especially considering loose garments.

% of how they compare to marker-based MoCap systems, especially in loose garments.
%Marker-less MoCap takes a different approach and utilizes DL models trained on annotated datasets to directly infer the 3D pose of individuals from input images or videos. By leveraging DL, the models learn to map semantic information to pose without explicit markers.
%Numerous marker-less DL MoCap models have been proposed for different types of motions, with marker-based MoCap systems used as the ground truth \cite{sigal2021human, chatzis2020comprehensive,gamra2021review}.
% There is extensive research on different marker-less DL MoCap models \cite{sigal2021human, chatzis2020comprehensive,gamra2021review} for various types of motions using marker-based MoCap as the ground truth.
% However, most of these studies do not consider that loose garments may compromise the pose from marker-based MoCap. 

Several studies comparing marker-based and marker-less MoCap in applications such as controlling an endoscopic instrument \cite{reilink20133d}, baseball pitching biomechanics \cite{fleisig2022comparison}, gait analysis \cite{kanko2021concurrent}, and clinical usability \cite{Ancans2021Wearable} have found that while marker-based MoCap generally exhibits slightly higher accuracy, marker-less systems have the potential to serve as a viable alternative, especially in clinical settings where patient comfort and ease of use are crucial factors \cite{nakano2020evaluation}. 

% Studies to compare marker-based and marker-less pose estimation has been conducted in the past using various application as the basis of comparison like controlling an endoscopic instrument\cite{reilink20133d}, baseball pitching biomechanics \cite{fleisig2022comparison}, for gait analysis \cite{kanko2021concurrent}, and clinical usability  \cite{Ancans2021Wearable}. 
\begin{figure*}[!t]
  \centering
  \includegraphics[width=\linewidth]{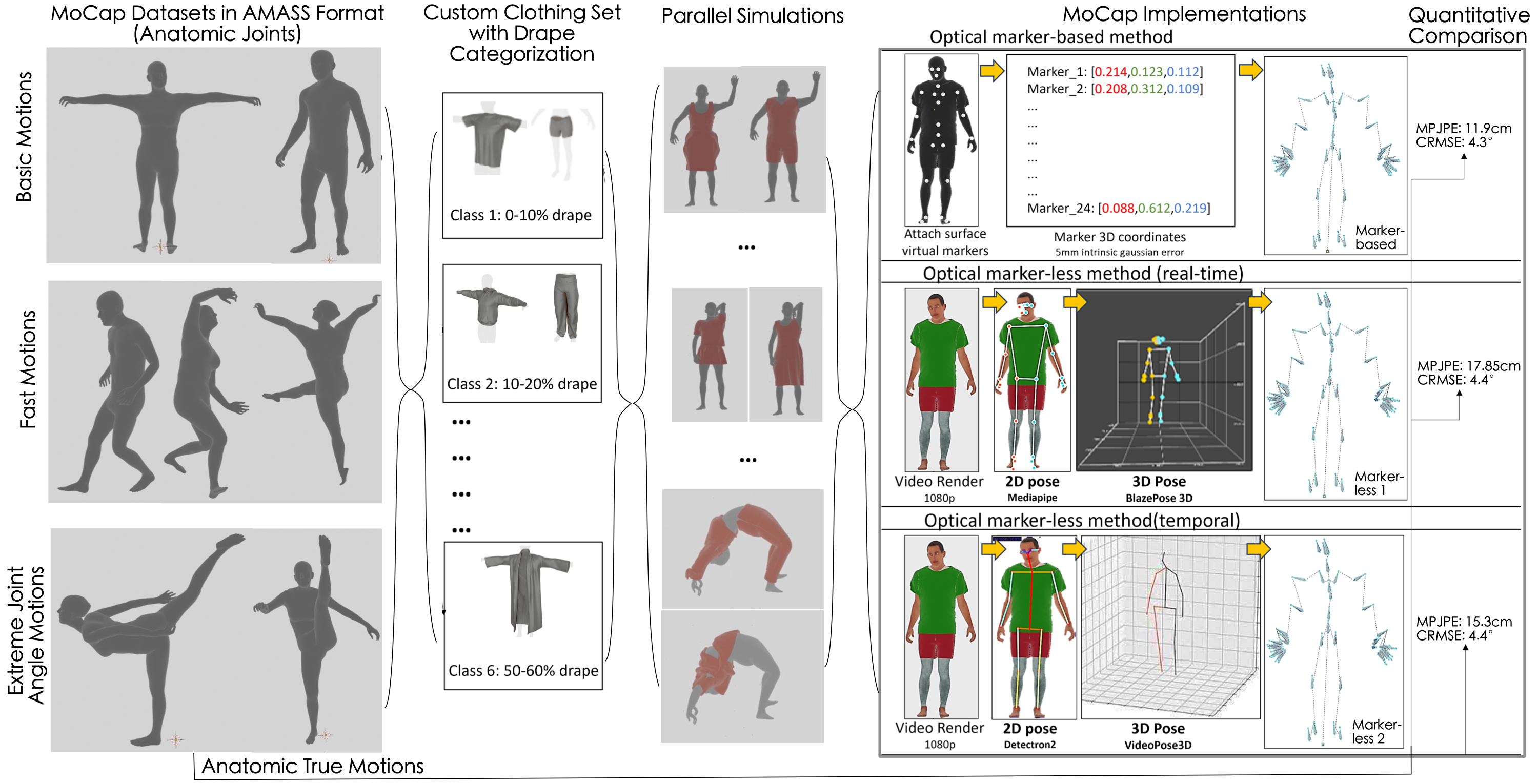}
  \vspace{-20pt}
  \caption{Overall pipeline of DMCB depicting pose estimation and calculation of MPJPE and CRMSE for all MoCap methods for different garment classes using a single motion data sequence.}
  \vspace{-10pt}
  \Description{ToDo}
  \label{fig:pipeline}
\end{figure*}
%Most of these studies have pointed out that although marker-based MoCap has slightly higher accuracy \cite{nakano2020evaluation}; marker-less systems have the potential to be a viable alternative, particularly for clinical settings where patient comfort and ease of use are essential factors. 
%Because of the absence of ground truth, these studies focus on complexity, ease of use, and overall performance rather than quantitative precision comparison. 
These studies prioritize complexity, ease of use, and overall performance rather than quantitative precision comparison due to the absence of anatomic motion reference, and no evaluation is done considering loose garments to the level of casual apparel.
However, to conduct such a comparison, especially with loose garments, there are two requirements that cannot be achieved in reality:
\begin{enumerate}
    \item Anatomic true motion under the garment and skin is required to quantitatively compare MoCap methods, which cannot be captured in reality through non-invasive methods. %, because even marker-based MoCap uses biomechanical approximation from surface markers.
    \item The exact motion sequences need to be reproduced precisely in scenarios with individuals of different body shapes wearing different garments.
\end{enumerate}
% (1) The anatomical true motion underneath the garment and skin is required to quantitatively compare different MoCap methods, which is unknown in the real world because even marker-based MoCap uses biomechanical approximation from surface markers.
% (2)  The exact motion sequences need to be reproduced precisely in multiple scenarios with persons of different body shapes wearing different garments.
Largely due to these challenges, existing quantitative reviews of marker-less methods use marker-based MoCap as reference \cite{wang2021deep}, which itself has substantial error from the anatomic joints due to the biomechanical approximation.
To address these issues, we leverage 3D physics-based simulation to benchmark the impossible.
We use real-world captured motion datasets to generate the inputs required for marker-based and marker-less MoCap methods, thus quantitatively comparing them to the common anatomic true motion. 
In particular, we make the following contributions: 
\begin{enumerate}
    \item A garment and soft body physics simulation evaluate marker-based and marker-less MoCap performances while persons with different body types repeat the same motions wearing different garments in terms of drape. Using real-world captured motion datasets to generate the input required for both MoCap methods, we quantitatively compare them to the common anatomical true motion.
    \item Our benchmark involves diverse varieties of motion types and garment drape levels which, together with a holistic comparison, can assist practitioners in choosing the optimal MoCap for wearable experiment ground truth for their specific applications balancing aspects such as garment designs, types of motion, cost and time overhead, and precision.
\end{enumerate}
% (1) A garment and soft body physics simulation evaluates marker-based and marker-less MoCap performances while persons with different body types repeat the same motions wearing different garments in terms of drape. Using real-world captured motion datasets to generate the input required for both MoCap methods, we quantitatively compare them to the common anatomical true motion. %, which is impossible through real-world captures. %. This solves the problem of needing to have ground truth pose in a real dataset. Reality check through 3D physics-based simulation: w

% (2) We quantitatively compared the MoCap methods in diverse varieties of motion types and garment drape levels. Together with a holistic comparison, our benchmark can assist practitioners in choosing the optimal MoCap for wearable experiment ground truth for their specific applications balancing aspects such as garment designs, types of motion, cost and time overhead, and precision.
% by three motion types (daily activity, fast, extreme joint) and six quantified drape levels. 
%The benchmark can be adapted to new motions and garments not included in our benchmark datasets. %We present a framework to translate MoCap data into videos of virtual persons wearing simulated clothes embedded with optical markers.
%\vspace{-8pt}

%\section{Background}
\vspace{-3pt}
% \vspace{-8pt}
\section{Proposed Method}
\vspace{-2pt}
The overall framework of our benchmark methodology is shown in \cref{fig:pipeline}.
By leveraging 3D physics simulation, we solve the reality challenge that the exact motion cannot be perfectly reproduced to establish quantitative comparisons of different scenarios.% with different MoCap methods.

\vspace{-8pt}
\subsection{DrapeMoCapBench Pipeline}

The simulation pipeline strictly adheres to reality, as the inputs to all MoCap methods are true to their specifications: 3D surface marker locations for marker-based kinematic methods and 1080p image sequences for marker-less vision models.

\vspace{-5pt}
\subsubsection{3D physics simulation}
With Blender3D \cite{blender}, motion sequences from \cref{sec:mocap_dataset} were converted to volumetric human bodies of different builds with the help of SMPL-X blender addon \cite{pavlakos2019expressive}, then dressed in garments described in \cref{sec:clothes}.
All simulated garments are assigned cloth properties equivalent to that of woven cotton (un-stretchable) with vertex mass of 0.05 kg, stiffness tension and compression of 15, stiffness and damping bending of 0.5, damping tension, compression, and shear 5, and stiffness shear of 10.
Doubled layered cloth mesh and improved body-cloth collision provided in Simplycloth \cite{simplycloth} along with soft tissue dynamics over captured skeletal motions using Mosh++\cite{loper2014mosh} enabled us to introduce realistic deformation of the garments over volumetric human models performing dynamic activities while having minimal artifacts.
Then inputs for optical MoCap were derived from the 3D scenes of parallel simulations of the same underlying motion as described in \cref{sec:mocap_methods}.

\vspace{-5pt}
\subsubsection{Motion Source Dataset}
\label{sec:mocap_dataset}
We used the AMASS framework \cite{mahmood2019amass} for converting MoCap data from various sources and formats to a standardized format based on the SMPL \cite{SMPL:2015} body model, a 3D model that accurately represents the human body. 
We considered three types of motion. First, the basic motions, such as walking and interaction, are derived from the HumanEva \cite{sigal2010humaneva} and TotalCapture \cite{trumble2017total} dataset containing 20 minutes of motion sequences. % 30 samples of total of 36210 frames. 
Second, the fast motions, including sports, dancing, etc., are derived from DanceDB \cite{dance_db} and Totalcapture containing 40 minutes of motion. %33 samples of a total of 70985 frames. 
Finally, the motions with extreme joint bending, like Yoga and gymnastics, are derived from PosePrior \cite{akhter2015pose}, and PresSim \cite{ray2023pressim} dataset containing 37 minutes of motion. % 36 samples having a total of 66560 frames.

\begin{figure}[!t]
  \centering
  \includegraphics[width=\linewidth]{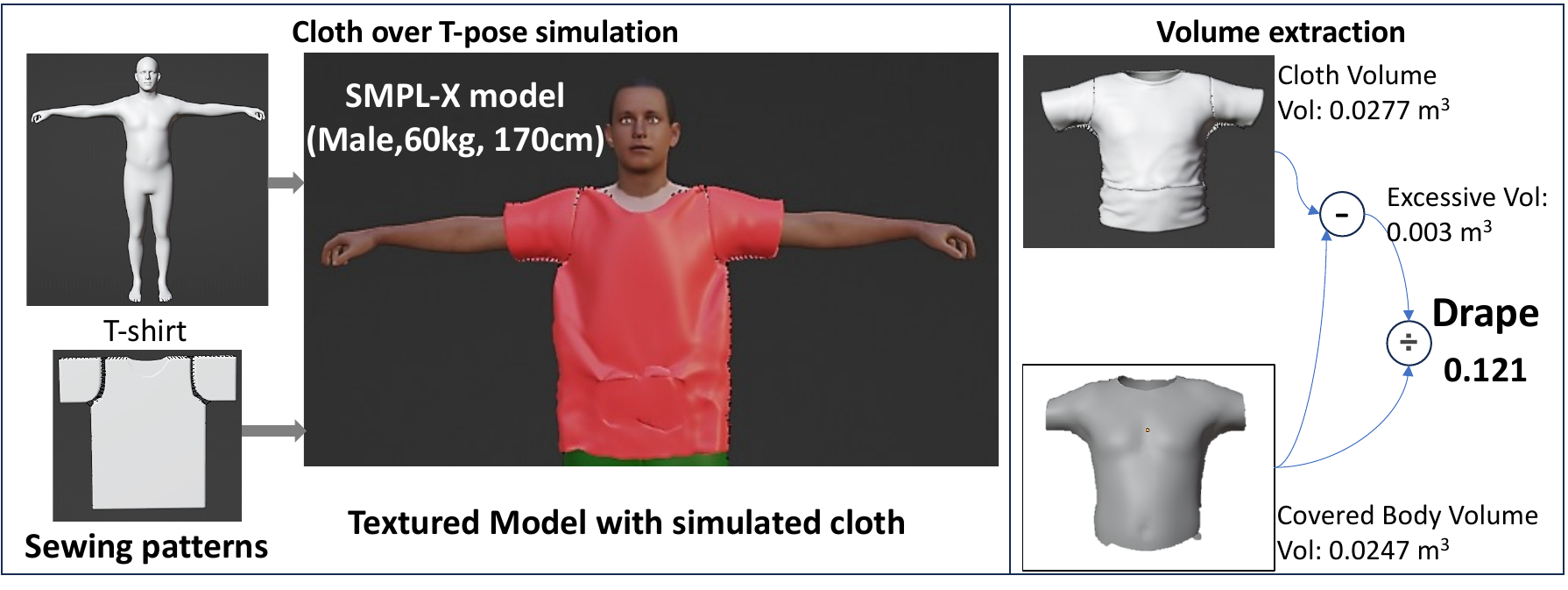}
  \vspace{-20pt}
  \caption{Quantifying drape for t-shirt and trousers.}
  \vspace{-15pt}
  \Description{ToDo}
  \label{fig:overview}
\end{figure}

%\begin{table}
%\footnotesize
%\centering
%\caption{Drape class categorization for different cloth, gender, build combinations.}
%\vspace{-10pt}
%\label{tab:draping}
%\begin{tabular}{ccccc}
%\toprule
%\multicolumn{2}{c}{Drape Classes} & Cloth & \multicolumn{2}{c}{Drape} \\
%\cmidrule{4-5}
%Male & Female & &  Male & Female\\
%\midrule
%1-1-1 & 1-1-1 & Sleeveless & 0.10-0.08-0.07 & 0.09-0.08-0.07\\
%2-1-1 & 2-1-1 & T-Shirt & 0.12-0.09-0.07 & 0.11-0.10-0.07\\
%3-2-1 & 2-2-1 & Shorts & 0.21-0.14-0.10 & 0.17-0.13-0.09\\
%3-2-2 & 2-2-1 & Skirt & 0.24-0.17-0.12 & 0.19-0.14-0.10\\
%3-2-2 & 2-2-2 & Shirt & 0.25-0.21-0.18 & 0.19-0.18-0.16\\
%3-2-2 & 3-2-2 & Dress & 0.21-0.19-0.16 & 0.21-0.19-0.17\\
%3-3-3 & 3-3-3 &Trousers & 0.28-0.25-0.23 & 0.29-0.26-0.24\\
%3-3-3 & 4-3-3 & Jacket & 0.29-0.27-0.26 & 0.31-0.28-0.27\\
%4-3-3 & 4-4-3 & Hoodie & 0.35-0.29-0.26 & 0.36-0.31-0.28\\
%4-4-3 & 4-4-3 & Cardigan & 0.33-0.31-0.29 & 0.32-0.31-0.28\\
%5-4-4 & 5-4-4 & Cargo & 0.41-0.36-0.33 & 0.41-0.38-0.35\\
%6-5-5 & 5-5-5 & Robes & 0.51-0.46-0.43 & 0.48-0.45-0.44\\
%6-5-5 & 6-5-5 & Trench Coat & 0.53-0.50-0.48 & 0.53-0.50-0.47\\
%\bottomrule
%\multicolumn{5}{c}
%{*For each 'male' and 'female', we separate by small - average - heavy build.}\\
%\end{tabular}
%\vspace{-10pt}
%\end{table}
\vspace{-5pt}
\subsubsection{Quantifying drape of loose garments}
\label{sec:clothes}
\begin{figure}[!t]
  \centering
  %\vspace{-10pt}
  \includegraphics[width=\linewidth]{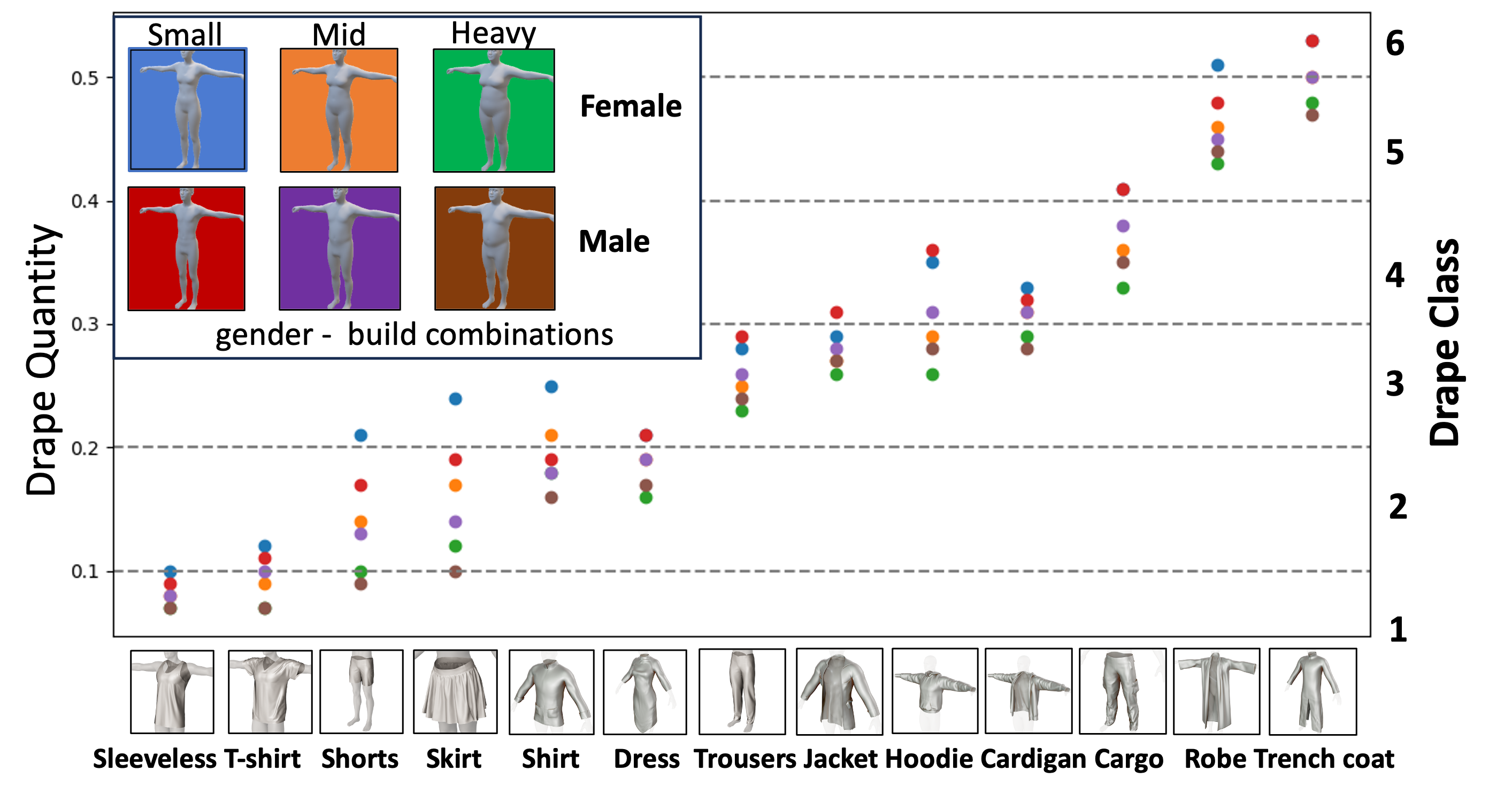}
  \vspace{-20pt}
  \caption{Drape class for different garments, gender \& builds.}
  \vspace{-15pt}
  \Description{ToDo}
  \label{fig:drape}
\end{figure}

We used 3D assets for a broad selection of apparel from commonly available categories for both genders using the Simplycloth\cite{simplycloth} plugin with garments ranging from skin-tight (minimal drape) to very loose (maximal drape). They are assigned into one of six classes based on the drape percentage, as depicted in \cref{fig:drape}. 
Drape amount is calculated as the percentage of the extra volume occupied by the garment as compared to another garment that fits the body perfectly and has the volume underlying the covered body as visualized in \cref{fig:overview}:
$
Drape = \frac{{Volume_{\text{garment}} - Volume_{\text{CoveredBody}}}}{{Volume_{\text{CoveredBody}}}}
$.
%It calculates the percentage of extra volume required by a garment as compared to another garment that fits the body perfectly.
For all garments except uni-cloths, a piece for the upper body and lower body for a particular build sharing combined drape class from 1 to 6 is selected to dress the SMPL body mesh.

\vspace{-5pt}
\subsection{MoCap methods}
\vspace{-3pt}
\label{sec:mocap_methods}
State-of-the-art marker-based and marker-less MoCap methods were implemented with the benchmark pipeline.
\vspace{-3pt}
\subsubsection{Marker-based}
%Although its very hard to say whether optical MoCap systems or inertial ones would perform better over loose garments, we choose to benchmark the former purely based on the advantage of better accuracy and possibility of absolute positioning. Inertial sensors get affected by noise and drift over time, which is rampant when used over loose garments.
Regardless of the marker principle, they return a 3D coordinate.
  % We assumed our markers to be active with an expected triangulation accuracy range of 5mm
A marker set of 24 marker pairs (48 markers) associated with the 24 joints from the SMPL skeleton is either attached over the garments or the skin, depending on whether their original position is covered by the garment in T-pose. 
%From each marker pairs $(x_1, y_1, z_1)$ and $(x_2, y_2, z_2)$  to calculate the position of each joint, we used
%$ Joint =  \frac{{x_1 + x_2}}{2}, \frac{{y_1 + y_2}}{2}, \frac{{z_1 + z_2}}{2} $.
This represents the optimal real-world marker placement, and 24 marker pairs are sufficient, as the simulation retrieves the coordinates directly without accounting for occlusion and triangulation from multiple cameras.
5 mm error was added as Gaussian noise to the coordinate according to the best-performing solutions \cite{merriaux2017study,maletsky2007accuracy,van2018accuracy}.
%The extracted 3D marker position is then normalized to have a range from -0.1 to 0.1.
%The origin of the skeleton is moved to the pelvic joint to match the output of the marker-less MoCap before being translated back 
The surface markers are converted to Bio-Vision Hierarchy (BVH) MoCap files having SMPL skeleton hierarchy using forward kinematics to approximate joints' absolute position and angle.
\vspace{-5pt}
\subsubsection{Marker-less}
Two marker-less models were considered: a temporal semi-supervised 3d pose estimation model VideoPose3D \cite{pavllo20193d} and a lightweight real-time 3d pose estimation model BlazePose3D \cite{bazarevsky2020blazepose}.
Videos (1920$\times$1080) were rendered from the simulation scene, then fed into Detectron2\cite{wu2019detectron2} + VideoPose3D or BlazePose3D to extract multi-joint poses relative to the video frame. They are then rescaled to the original size of the body (170cm height) and converted to BVH files.
%to extract the 17-joint 3D poses, which are mapped back to 24 joint SMPL poses using correspondence mapping and transformation estimation and then converted to BVH MoCap files containing SMPL skeleton hierarchy. % but having a range from -0.1 to 0.1 
%Similarly, we also calculated the BVH MoCap files from the 33-joint skeleton of BlazePose3D.

\vspace{-6pt}
\subsection{Evaluation Metrics}
\vspace{-3pt}
We consider two frequently used metrics in the MoCap field: absolute Mean Per Joint Position Error (MPJPE), which provides a quantitative measure of the accuracy of 3D joint positions and is more often used in animation, and Circular Root Mean Squared Error (CRMSE), that assesses the performance of pose joint angle estimation used primarily in sports and medical research on joint angles.
They are defined as follows: 
$ MPJPE = \frac{1}{n}\sum_{i=1}^{n} || \mathbf{P}_i - \mathbf{\hat{P}}_i || $
where $n$ is the number of joints, $\mathbf{P}_i$ is the ground truth position of the $i$-th joint, $\mathbf{\hat{P}}_i$ is the estimated position of the $i$-th joint, and $|| \cdot ||$ denotes the Euclidean distance. 
$
CRMSE = \sqrt{\frac{1}{N} \sum_{i=1}^{N} \left(1 - \cos(\theta_i - \hat{\theta_i})\right)}
$
Here, N represents the total number of joint angles, \(\theta_i\) represents the ground truth angle for the $i$-th joint, and \(\hat{\theta_i}\) represents the corresponding predicted angle.  
Due to the availability of comprehensive measurements about human models and garments obtained from the simulation, it is straightforward to calculate MPJPE using the 3D estimated joints in Euler space. On the other hand, the CRMSE involves estimating joint angles by applying forward kinematics and then computing the error.

\section{Results and Discussion}
\begin{figure}[!t]
  \centering
  \vspace{-10pt}
  \includegraphics[width=\linewidth]{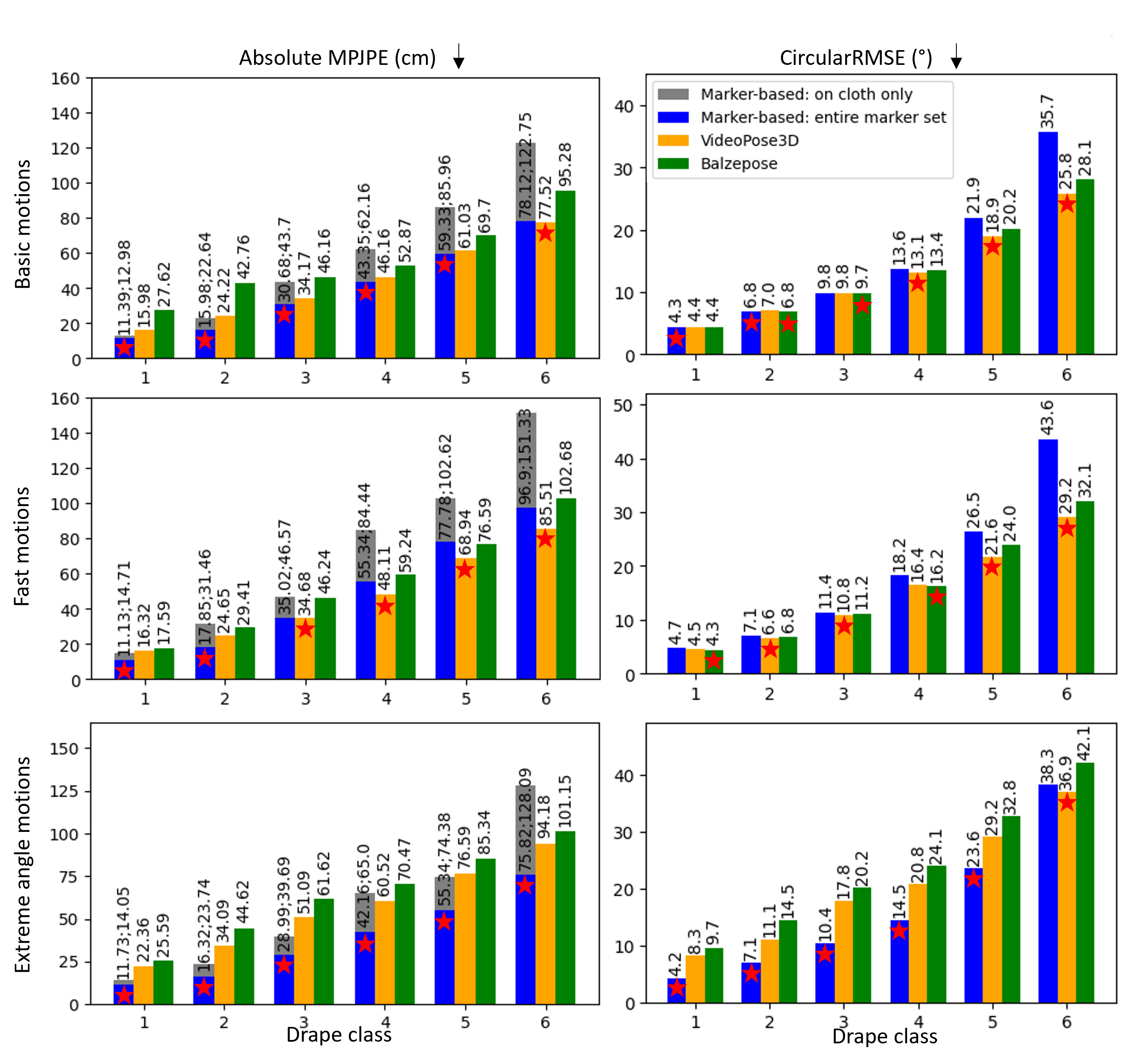}
  \vspace{-20pt}
  \caption{Benchmark results of marker-based \& marker-less MoCap methods in different drape \& motion combinations.}
  \vspace{-15pt}
  \Description{ToDo}
  \label{fig:benchmark}
\end{figure}

\begin{figure}[!t]
  \centering
    \vspace{-10pt}
  \includegraphics[width=\linewidth]{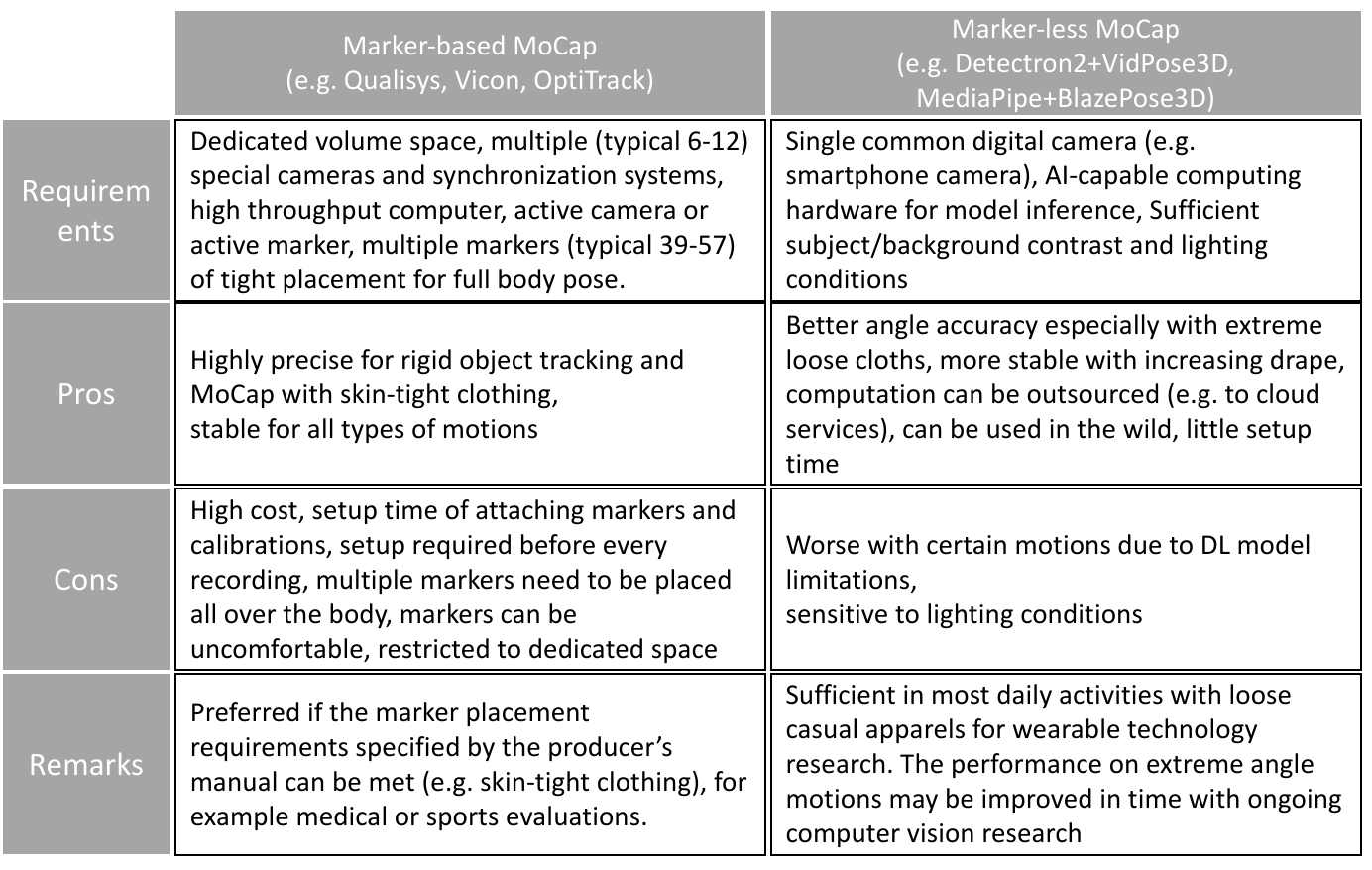}
  \vspace{-20pt}
  \caption{Holistic comparisons for two MoCap methods}
  \vspace{-15pt}
  \Description{ToDo}
  \label{fig:comp}
\end{figure}

\vspace{-3pt}
With the unclothed body, the MPJPE between marker-based and marker-less implementations of basic motion tested on the TotalCapture dataset is 4.7 cm, and extreme angle motions tested on the PosePrior dataset is 8.2 cm.
These results quantitatively align with the literature comparing marker-less models with marker-based mocap as reference \cite{kanazawa2018end, qiu2019cross, wang2021deep, ostrek2019existing}. 
The MPJPE and CRMSE between different MoCap implementations and the anatomic joints are detailed in \cref{fig:benchmark}. We calculated MPJPE for both markers placed on cloth only as well as the entire marker set for marker-based MoCap method. 
%Among the 18 combinations of 3 motion types and six drape levels, we can expect that in 16 cases, both marker-based and marker-less methods will provide less than 1cm joint position error. 
%The differences between marker-based and marker-less methods are generally within 0.1cm MPJPE.
%Exceptions are the extremely loose (drape class 6) with fast or extreme joint motions, which still see a maximum 1.1cm error from both approaches. 
The minimum joint-position error is unsurprisingly from drape class 1 with marker-based methods, which is still >10 cm.
Such comparison has only been possible before with our simulation pipeline, as the anatomic joint coordinates cannot be derived in reality with non-invasive superficial methods like surface markers or video analysis as explained in \cref{sec:intro}.
The term `looseness' is highly subjective, and its interpretation depends on the relative volume of the garment as compared to the wearer. 
%For instance, a shirt that is normally sized may be perceived as loose if worn by a slender individual. 
To account for this variability, we employed a quantitative measurement of drape and organized our findings into drape classes. 
Everyday loose garments that effectively follow the wearer's body motion typically belong to drape class 2 or 3. 
In this range, either marker-based or marker-less gives 15cm to 35cm MPJPE and 6° to 11° CRMSE.
Absolute MPJPE is susceptible to joint hierarchy alignment, including shifting and rotation errors, while CRMSE is affected by the relative angle of bone in terms of its parent. As expected bio-mechanical constraints the MPJPE overall marker set as compared to where it is calculated for markers placed only over the garment in marker-based MoCap.
However, there are extreme cases, such as robes and trench coats, where the garments exhibit limited adherence to the wearer's motion.
All methods see gradually increasing errors with more drape of the garment.%; however, the rate of degradation of marker-less methods is slightly less for basic motions and fast motions that can be visualized by calculating the difference to skin-tight baselines of each method to quantify the degradation of MoCap accuracy over draping which shows slightly less range of variation for marker-less methods irrespective of motion type.
Marker-less methods (especially VideoPose3D) are more stable as the drape increases, which might be attributed to the fact that they work on detecting semantic segments of body parts.
DL-based marker-less models are expected to perform better on basic and fast motions than on extreme joint angles since they are primarily trained on datasets consisting of the first two.
On the other hand, the marker-based method does not have such limitations, as it uses forward kinematics with biomechanical constraints from the markers. % rather than a trained DL model.
Apart from the quantitative benchmark results, we also compare both methods holistically as listed in \cref{fig:comp}.

In our simulated benchmark, we aimed to maintain realism as much as possible; however, there are certain limitations that could be addressed in future research. 
For the MoCap implementations, the marker-based method excluded occlusions in real-world triangulation and human setup errors as marker positions are retrieved directly from the simulation; and marker-less methods did not consider optical confusions, especially between the subject and background, which may lead to sub-optimal image quality.
However, these aspects are also challenges in real-world experiments and are often actively eliminated by practitioners.
Our simulation only considered one standard cloth material; the influence of different materials on how the garment and markers react to various motions could be included. Draping is significantly affected by cloth material as it determines the weight, stretch, and rigidity which influences how the fabric hangs, folds, and holds a shape when draped over a 3D form. 
To further enhance realism, future work should incorporate support for external collisions, such as garments interacting with the floor. The variety of clothing options and motion types could also be expanded. 
It is also crucial to introduce support for other modalities in our benchmarking multi-camera and RGBD marker-less MoCap methods, which could provide even better precision with still less cost than marker-based MoCap.

\vspace{-3pt}
\section{Conclusions and Outlook}
\vspace{-3pt}
We propose the DrapeMoCapBench as a benchmark methodology based on 3D physics simulations to compare marker-based and marker-less MoCap systems, mainly when dealing with loose garments. 
The simulation allows us to benchmark what is impossible in reality - quantitatively comparing different scenarios of precisely reproduced motions against the anatomic true motion under draped garments and body skin.
It incorporates physics-based simulation of the human body and garment models with real-world motion datasets, generating input data for marker-based and marker-less MoCap methods. 
Through the benchmark, we provide a comprehensive comparison of the MoCap methods with a benchmark table that quantifies precision for different motion types and levels of garment drape and holistic considerations. 
Our findings indicate that, while in line with the literature for skin-tight clothing, both marker-based and marker-less MoCap suffer significant performance loss in casual loose garments like shirts.
For daily activities (basic and fast motions) involving casual loose garments, marker-less MoCap slightly outperforms marker-based MoCap.
Considering the low cost of marker-less methods, they could be a preferable choice of reference for wearable studies.
DMCB can be a valuable resource for wearable practitioners seeking to select the most suitable MoCap method for their own applications, considering scenario-specific precision and holistic factors. 
The marker-less methods are closer to the marker-based MoCap than the anatomic joints in terms of MPJPE, which could be explained by the fact that the DL models were trained mostly using marker-based MoCap as the ground truth.
DMCB can also be used in future work as a data augmentation tool to improve vision-based pose estimation models.
Wearable developers can consider the specific errors we identified for each motion and garment type rather than relying solely on marker tracking errors, which through our findings, do not accurately represent the pose estimation error often not specified explicitly by the MoCap system.
Furthermore, the DMCB pipeline can be adapted to evaluate the expected MoCap performance for new garment designs, allowing for tailored assessments of different garment pieces.
\newpage
\begin{acks}
%To Robert, for the bagels and explaining CMYK and color spaces.
The research reported in this paper was supported by the BMBF (German Federal Ministry of Education and Research) in the project VidGenSense (01IW21003). It was also funded by Carl-Zeiss Stiftung under the Sustainable Embedded AI project (P2021-02-009).
\end{acks}

\bibliographystyle{ACM-Reference-Format}
\bibliography{sample-base}


\begin{thebibliography}{50}

%%% the Web address.

\ifx \showCODEN    \undefined \def \showCODEN     #1{\unskip}     \fi
\ifx \showDOI      \undefined \def \showDOI       #1{#1}\fi
\ifx \showISBNx    \undefined \def \showISBNx     #1{\unskip}     \fi
\ifx \showISBNxiii \undefined \def \showISBNxiii  #1{\unskip}     \fi
\ifx \showISSN     \undefined \def \showISSN      #1{\unskip}     \fi
\ifx \showLCCN     \undefined \def \showLCCN      #1{\unskip}     \fi
\ifx \shownote     \undefined \def \shownote      #1{#1}          \fi
\ifx \showarticletitle \undefined \def \showarticletitle #1{#1}   \fi
\ifx \showURL      \undefined \def \showURL       {\relax}        \fi
% invisible to TeX
\providecommand\bibfield[2]{#2}
\providecommand\bibinfo[2]{#2}
\providecommand\natexlab[1]{#1}
\providecommand\showeprint[2][]{arXiv:#2}

\bibitem[sim(2022)]%
        {simplycloth}
 \bibinfo{year}{2022}\natexlab{}.
\newblock \bibinfo{title}{SimplyCloth}.
\newblock \bibinfo{howpublished}{Retrieved May 23, 2023, from
  \url{https://blendermarket.com/products/simply-cloth}}.
\newblock


\bibitem[ble(2023)]%
        {blender}
 \bibinfo{year}{2023}\natexlab{}.
\newblock \bibinfo{title}{Blender}.
\newblock \bibinfo{howpublished}{Computer software}.
\newblock
\urldef\tempurl%
\url{https://www.blender.org}
\showURL{%
\tempurl}


\bibitem[Akhter and Black(2015)]%
        {akhter2015pose}
\bibfield{author}{\bibinfo{person}{Ijaz Akhter} {and}
  \bibinfo{person}{Michael~J Black}.} \bibinfo{year}{2015}\natexlab{}.
\newblock \showarticletitle{Pose-conditioned joint angle limits for 3D human
  pose reconstruction}. In \bibinfo{booktitle}{\emph{Proceedings of the IEEE
  conference on computer vision and pattern recognition}}.
  \bibinfo{pages}{1446--1455}.
\newblock


\bibitem[Ancans(2021)]%
        {Ancans2021Wearable}
\bibfield{author}{\bibinfo{person}{A. Ancans}.}
  \bibinfo{year}{2021}\natexlab{}.
\newblock \showarticletitle{Wearable Sensor Clothing for Body Movement
  Measurement during Physical Activities in Healthcare}.
\newblock \bibinfo{journal}{\emph{Sensors (Basel, Switzerland)}}
  (\bibinfo{year}{2021}).
\newblock


\bibitem[Barca et~al\mbox{.}(2006)]%
        {barca2006new}
\bibfield{author}{\bibinfo{person}{Jan~Carlo Barca}, \bibinfo{person}{Grace
  Rumantir}, {and} \bibinfo{person}{Raymond~Koon Li}.}
  \bibinfo{year}{2006}\natexlab{}.
\newblock \showarticletitle{A new illuminated contour-based marker system for
  optical motion capture}. In \bibinfo{booktitle}{\emph{2006 Innovations in
  Information Technology}}. IEEE, \bibinfo{pages}{1--5}.
\newblock


\bibitem[Bazarevsky et~al\mbox{.}(2020)]%
        {bazarevsky2020blazepose}
\bibfield{author}{\bibinfo{person}{Valentin Bazarevsky}, \bibinfo{person}{Ivan
  Grishchenko}, \bibinfo{person}{Karthik Raveendran}, \bibinfo{person}{Tyler
  Zhu}, \bibinfo{person}{Fan Zhang}, {and} \bibinfo{person}{Matthias
  Grundmann}.} \bibinfo{year}{2020}\natexlab{}.
\newblock \showarticletitle{Blazepose: On-device real-time body pose tracking}.
\newblock \bibinfo{journal}{\emph{arXiv preprint arXiv:2006.10204}}
  (\bibinfo{year}{2020}).
\newblock


\bibitem[Behera et~al\mbox{.}(2020)]%
        {behera2020deep}
\bibfield{author}{\bibinfo{person}{Ardhendu Behera}, \bibinfo{person}{Zachary
  Wharton}, \bibinfo{person}{Alexander Keidel}, {and}
  \bibinfo{person}{Bappaditya Debnath}.} \bibinfo{year}{2020}\natexlab{}.
\newblock \showarticletitle{Deep cnn, body pose, and body-object interaction
  features for drivers’ activity monitoring}.
\newblock \bibinfo{journal}{\emph{IEEE Transactions on Intelligent
  Transportation Systems}} \bibinfo{volume}{23}, \bibinfo{number}{3}
  (\bibinfo{year}{2020}), \bibinfo{pages}{2874--2881}.
\newblock


\bibitem[Bello et~al\mbox{.}(2021)]%
        {bello2021mocapaci}
\bibfield{author}{\bibinfo{person}{Hymalai Bello}, \bibinfo{person}{Bo Zhou},
  \bibinfo{person}{Sungho Suh}, {and} \bibinfo{person}{Paul Lukowicz}.}
  \bibinfo{year}{2021}\natexlab{}.
\newblock \showarticletitle{Mocapaci: Posture and gesture detection in loose
  garments using textile cables as capacitive antennas}. In
  \bibinfo{booktitle}{\emph{Proceedings of the 2021 ACM International Symposium
  on Wearable Computers}}. \bibinfo{pages}{78--83}.
\newblock


\bibitem[Chatzis et~al\mbox{.}(2020)]%
        {chatzis2020comprehensive}
\bibfield{author}{\bibinfo{person}{Theocharis Chatzis},
  \bibinfo{person}{Andreas Stergioulas}, \bibinfo{person}{Dimitrios
  Konstantinidis}, \bibinfo{person}{Kosmas Dimitropoulos}, {and}
  \bibinfo{person}{Petros Daras}.} \bibinfo{year}{2020}\natexlab{}.
\newblock \showarticletitle{A comprehensive study on deep learning-based 3D
  hand pose estimation methods}.
\newblock \bibinfo{journal}{\emph{Applied Sciences}} \bibinfo{volume}{10},
  \bibinfo{number}{19} (\bibinfo{year}{2020}), \bibinfo{pages}{6850}.
\newblock


\bibitem[Fleisig et~al\mbox{.}(2022)]%
        {fleisig2022comparison}
\bibfield{author}{\bibinfo{person}{Glenn~S Fleisig},
  \bibinfo{person}{Jonathan~S Slowik}, \bibinfo{person}{Derek Wassom},
  \bibinfo{person}{Yuki Yanagita}, \bibinfo{person}{Jasper Bishop}, {and}
  \bibinfo{person}{Alek Diffendaffer}.} \bibinfo{year}{2022}\natexlab{}.
\newblock \showarticletitle{Comparison of marker-less and marker-based motion
  capture for baseball pitching kinematics}.
\newblock \bibinfo{journal}{\emph{Sports Biomechanics}} (\bibinfo{year}{2022}),
  \bibinfo{pages}{1--10}.
\newblock


\bibitem[Fleron et~al\mbox{.}(2019)]%
        {fleron2019accuracy}
\bibfield{author}{\bibinfo{person}{Martin~Kokholm Fleron},
  \bibinfo{person}{Niels Christian~Hauerbach Ubbesen},
  \bibinfo{person}{Francesco Battistella}, \bibinfo{person}{David~Leandro
  Dejtiar}, {and} \bibinfo{person}{Anderson~Souza Oliveira}.}
  \bibinfo{year}{2019}\natexlab{}.
\newblock \showarticletitle{Accuracy between optical and inertial motion
  capture systems for assessing trunk speed during preferred gait and
  transition periods}.
\newblock \bibinfo{journal}{\emph{Sports biomechanics}} \bibinfo{volume}{18},
  \bibinfo{number}{4} (\bibinfo{year}{2019}), \bibinfo{pages}{366--377}.
\newblock


\bibitem[Gamra and Akhloufi(2021)]%
        {gamra2021review}
\bibfield{author}{\bibinfo{person}{Miniar~Ben Gamra} {and}
  \bibinfo{person}{Moulay~A Akhloufi}.} \bibinfo{year}{2021}\natexlab{}.
\newblock \showarticletitle{A review of deep learning techniques for 2D and 3D
  human pose estimation}.
\newblock \bibinfo{journal}{\emph{Image and Vision Computing}}
  \bibinfo{volume}{114} (\bibinfo{year}{2021}), \bibinfo{pages}{104282}.
\newblock


\bibitem[Gong et~al\mbox{.}(2021)]%
        {gong2021robust}
\bibfield{author}{\bibinfo{person}{Jian Gong}, \bibinfo{person}{Xinyu Zhang},
  \bibinfo{person}{Yuanjun Huang}, \bibinfo{person}{Ju Ren}, {and}
  \bibinfo{person}{Yaoxue Zhang}.} \bibinfo{year}{2021}\natexlab{}.
\newblock \showarticletitle{Robust inertial motion tracking through deep sensor
  fusion across smart earbuds and smartphone}.
\newblock \bibinfo{journal}{\emph{Proceedings of the ACM on Interactive,
  Mobile, Wearable and Ubiquitous Technologies}} \bibinfo{volume}{5},
  \bibinfo{number}{2} (\bibinfo{year}{2021}), \bibinfo{pages}{1--26}.
\newblock


\bibitem[Groen et~al\mbox{.}(2012)]%
        {groen2012sensitivity}
\bibfield{author}{\bibinfo{person}{Brenda~E Groen}, \bibinfo{person}{Marjolein
  Geurts}, \bibinfo{person}{Bart Nienhuis}, {and} \bibinfo{person}{Jacques
  Duysens}.} \bibinfo{year}{2012}\natexlab{}.
\newblock \showarticletitle{Sensitivity of the OLGA and VCM models to erroneous
  marker placement: Effects on 3D-gait kinematics}.
\newblock \bibinfo{journal}{\emph{Gait \& posture}} \bibinfo{volume}{35},
  \bibinfo{number}{3} (\bibinfo{year}{2012}), \bibinfo{pages}{517--521}.
\newblock


\bibitem[Jansen et~al\mbox{.}(2007)]%
        {jansen20073d}
\bibfield{author}{\bibinfo{person}{Bart Jansen}, \bibinfo{person}{Frederik
  Temmermans}, {and} \bibinfo{person}{Rudi Deklerck}.}
  \bibinfo{year}{2007}\natexlab{}.
\newblock \showarticletitle{3D human pose recognition for home monitoring of
  elderly}. In \bibinfo{booktitle}{\emph{2007 29th Annual International
  Conference of the IEEE Engineering in Medicine and Biology Society}}. IEEE,
  \bibinfo{pages}{4049--4051}.
\newblock


\bibitem[Jiang et~al\mbox{.}(2022)]%
        {jiang2022transformer}
\bibfield{author}{\bibinfo{person}{Yifeng Jiang}, \bibinfo{person}{Yuting Ye},
  \bibinfo{person}{Deepak Gopinath}, \bibinfo{person}{Jungdam Won},
  \bibinfo{person}{Alexander~W Winkler}, {and} \bibinfo{person}{C~Karen Liu}.}
  \bibinfo{year}{2022}\natexlab{}.
\newblock \showarticletitle{Transformer Inertial Poser: Real-time Human Motion
  Reconstruction from Sparse IMUs with Simultaneous Terrain Generation}. In
  \bibinfo{booktitle}{\emph{SIGGRAPH Asia 2022 Conference Papers}}.
  \bibinfo{pages}{1--9}.
\newblock


\bibitem[Jin et~al\mbox{.}(2018)]%
        {jin2018towards}
\bibfield{author}{\bibinfo{person}{Haojian Jin}, \bibinfo{person}{Zhijian
  Yang}, \bibinfo{person}{Swarun Kumar}, {and} \bibinfo{person}{Jason~I Hong}.}
  \bibinfo{year}{2018}\natexlab{}.
\newblock \showarticletitle{Towards wearable everyday body-frame tracking using
  passive RFIDs}.
\newblock \bibinfo{journal}{\emph{Proceedings of the ACM on Interactive,
  Mobile, Wearable and Ubiquitous Technologies}} \bibinfo{volume}{1},
  \bibinfo{number}{4} (\bibinfo{year}{2018}), \bibinfo{pages}{1--23}.
\newblock


\bibitem[Kanazawa et~al\mbox{.}(2018)]%
        {kanazawa2018end}
\bibfield{author}{\bibinfo{person}{Angjoo Kanazawa}, \bibinfo{person}{Michael~J
  Black}, \bibinfo{person}{David~W Jacobs}, {and} \bibinfo{person}{Jitendra
  Malik}.} \bibinfo{year}{2018}\natexlab{}.
\newblock \showarticletitle{End-to-end recovery of human shape and pose}. In
  \bibinfo{booktitle}{\emph{Proceedings of the IEEE conference on computer
  vision and pattern recognition}}. \bibinfo{pages}{7122--7131}.
\newblock


\bibitem[Kanko et~al\mbox{.}(2021)]%
        {kanko2021concurrent}
\bibfield{author}{\bibinfo{person}{Robert~M Kanko}, \bibinfo{person}{Elise~K
  Laende}, \bibinfo{person}{Elysia~M Davis}, \bibinfo{person}{W~Scott Selbie},
  {and} \bibinfo{person}{Kevin~J Deluzio}.} \bibinfo{year}{2021}\natexlab{}.
\newblock \showarticletitle{Concurrent assessment of gait kinematics using
  marker-based and markerless motion capture}.
\newblock \bibinfo{journal}{\emph{Journal of biomechanics}}
  \bibinfo{volume}{127} (\bibinfo{year}{2021}), \bibinfo{pages}{110665}.
\newblock


\bibitem[Lee and Yoo(2017)]%
        {lee2017low}
\bibfield{author}{\bibinfo{person}{Yeonkyung Lee} {and} \bibinfo{person}{Hoon
  Yoo}.} \bibinfo{year}{2017}\natexlab{}.
\newblock \showarticletitle{Low-cost 3D motion capture system using passive
  optical markers and monocular vision}.
\newblock \bibinfo{journal}{\emph{Optik}}  \bibinfo{volume}{130}
  (\bibinfo{year}{2017}), \bibinfo{pages}{1397--1407}.
\newblock


\bibitem[Liu et~al\mbox{.}(2019)]%
        {liu2019reconstructing}
\bibfield{author}{\bibinfo{person}{Ruibo Liu}, \bibinfo{person}{Qijia Shao},
  \bibinfo{person}{Siqi Wang}, \bibinfo{person}{Christina Ru},
  \bibinfo{person}{Devin Balkcom}, {and} \bibinfo{person}{Xia Zhou}.}
  \bibinfo{year}{2019}\natexlab{}.
\newblock \showarticletitle{Reconstructing human joint motion with
  computational fabrics}.
\newblock \bibinfo{journal}{\emph{Proceedings of the ACM on Interactive,
  Mobile, Wearable and Ubiquitous Technologies}} \bibinfo{volume}{3},
  \bibinfo{number}{1} (\bibinfo{year}{2019}), \bibinfo{pages}{1--26}.
\newblock


\bibitem[Liu(2020)]%
        {Liu2020A}
\bibfield{author}{\bibinfo{person}{Shiqiang Liu}.}
  \bibinfo{year}{2020}\natexlab{}.
\newblock \showarticletitle{A wearable motion capture device able to detect
  dynamic motion of human limbs}.
\newblock \bibinfo{journal}{\emph{Nature Communications}}
  (\bibinfo{year}{2020}).
\newblock


\bibitem[Loper et~al\mbox{.}(2014)]%
        {loper2014mosh}
\bibfield{author}{\bibinfo{person}{Matthew Loper}, \bibinfo{person}{Naureen
  Mahmood}, {and} \bibinfo{person}{Michael~J Black}.}
  \bibinfo{year}{2014}\natexlab{}.
\newblock \showarticletitle{MoSh: motion and shape capture from sparse
  markers.}
\newblock \bibinfo{journal}{\emph{ACM Trans. Graph.}} \bibinfo{volume}{33},
  \bibinfo{number}{6} (\bibinfo{year}{2014}), \bibinfo{pages}{220--1}.
\newblock


\bibitem[Loper et~al\mbox{.}(2015)]%
        {SMPL:2015}
\bibfield{author}{\bibinfo{person}{Matthew Loper}, \bibinfo{person}{Naureen
  Mahmood}, \bibinfo{person}{Javier Romero}, \bibinfo{person}{Gerard
  Pons-Moll}, {and} \bibinfo{person}{Michael~J. Black}.}
  \bibinfo{year}{2015}\natexlab{}.
\newblock \showarticletitle{{SMPL}: A Skinned Multi-Person Linear Model}.
\newblock \bibinfo{journal}{\emph{ACM Trans. Graphics (Proc. SIGGRAPH Asia)}}
  \bibinfo{volume}{34}, \bibinfo{number}{6} (\bibinfo{date}{Oct.}
  \bibinfo{year}{2015}), \bibinfo{pages}{248:1--248:16}.
\newblock


\bibitem[Mahmood et~al\mbox{.}(2019)]%
        {mahmood2019amass}
\bibfield{author}{\bibinfo{person}{Naureen Mahmood}, \bibinfo{person}{Nima
  Ghorbani}, \bibinfo{person}{Nikolaus~F Troje}, \bibinfo{person}{Gerard
  Pons-Moll}, {and} \bibinfo{person}{Michael~J Black}.}
  \bibinfo{year}{2019}\natexlab{}.
\newblock \showarticletitle{AMASS: Archive of motion capture as surface
  shapes}. In \bibinfo{booktitle}{\emph{Proceedings of the IEEE/CVF
  international conference on computer vision}}. \bibinfo{pages}{5442--5451}.
\newblock


\bibitem[Maletsky et~al\mbox{.}(2007)]%
        {maletsky2007accuracy}
\bibfield{author}{\bibinfo{person}{Lorin~P Maletsky}, \bibinfo{person}{Junyi
  Sun}, {and} \bibinfo{person}{Nicholas~A Morton}.}
  \bibinfo{year}{2007}\natexlab{}.
\newblock \showarticletitle{Accuracy of an optical active-marker system to
  track the relative motion of rigid bodies}.
\newblock \bibinfo{journal}{\emph{Journal of biomechanics}}
  \bibinfo{volume}{40}, \bibinfo{number}{3} (\bibinfo{year}{2007}),
  \bibinfo{pages}{682--685}.
\newblock


\bibitem[McAdams et~al\mbox{.}(2011)]%
        {mcadams2011wearable}
\bibfield{author}{\bibinfo{person}{Eric McAdams}, \bibinfo{person}{Asta
  Krupaviciute}, \bibinfo{person}{Claudine G{\'e}hin}, \bibinfo{person}{Etienne
  Grenier}, \bibinfo{person}{Bertrand Massot}, \bibinfo{person}{Andr{\'e}
  Dittmar}, \bibinfo{person}{Paul Rubel}, {and} \bibinfo{person}{Jocelyne
  Fayn}.} \bibinfo{year}{2011}\natexlab{}.
\newblock \showarticletitle{Wearable sensor systems: The challenges}. In
  \bibinfo{booktitle}{\emph{2011 Annual International Conference of the IEEE
  Engineering in Medicine and Biology Society}}. IEEE,
  \bibinfo{pages}{3648--3651}.
\newblock


\bibitem[McFadden et~al\mbox{.}(2020)]%
        {mcfadden2020sensitivity}
\bibfield{author}{\bibinfo{person}{Ciar{\'a}n McFadden},
  \bibinfo{person}{Katherine Daniels}, {and} \bibinfo{person}{Siobh{\'a}n
  Strike}.} \bibinfo{year}{2020}\natexlab{}.
\newblock \showarticletitle{The sensitivity of joint kinematics and kinetics to
  marker placement during a change of direction task}.
\newblock \bibinfo{journal}{\emph{Journal of Biomechanics}}
  \bibinfo{volume}{101} (\bibinfo{year}{2020}), \bibinfo{pages}{109635}.
\newblock


\bibitem[Merriaux et~al\mbox{.}(2017)]%
        {merriaux2017study}
\bibfield{author}{\bibinfo{person}{Pierre Merriaux}, \bibinfo{person}{Yohan
  Dupuis}, \bibinfo{person}{R{\'e}mi Boutteau}, \bibinfo{person}{Pascal
  Vasseur}, {and} \bibinfo{person}{Xavier Savatier}.}
  \bibinfo{year}{2017}\natexlab{}.
\newblock \showarticletitle{A study of vicon system positioning performance}.
\newblock \bibinfo{journal}{\emph{Sensors}} \bibinfo{volume}{17},
  \bibinfo{number}{7} (\bibinfo{year}{2017}), \bibinfo{pages}{1591}.
\newblock


\bibitem[Moon et~al\mbox{.}(2022)]%
        {moon2022imu2clip}
\bibfield{author}{\bibinfo{person}{Seungwhan Moon}, \bibinfo{person}{Andrea
  Madotto}, \bibinfo{person}{Zhaojiang Lin}, \bibinfo{person}{Alireza
  Dirafzoon}, \bibinfo{person}{Aparajita Saraf}, \bibinfo{person}{Amy Bearman},
  {and} \bibinfo{person}{Babak Damavandi}.} \bibinfo{year}{2022}\natexlab{}.
\newblock \showarticletitle{IMU2CLIP: Multimodal Contrastive Learning for IMU
  Motion Sensors from Egocentric Videos and Text}.
\newblock \bibinfo{journal}{\emph{arXiv preprint arXiv:2210.14395}}
  (\bibinfo{year}{2022}).
\newblock


\bibitem[Nakano et~al\mbox{.}(2020)]%
        {nakano2020evaluation}
\bibfield{author}{\bibinfo{person}{Nobuyasu Nakano}, \bibinfo{person}{Tetsuro
  Sakura}, \bibinfo{person}{Kazuhiro Ueda}, \bibinfo{person}{Leon Omura},
  \bibinfo{person}{Arata Kimura}, \bibinfo{person}{Yoichi Iino},
  \bibinfo{person}{Senshi Fukashiro}, {and} \bibinfo{person}{Shinsuke
  Yoshioka}.} \bibinfo{year}{2020}\natexlab{}.
\newblock \showarticletitle{Evaluation of 3D markerless motion capture accuracy
  using OpenPose with multiple video cameras}.
\newblock \bibinfo{journal}{\emph{Frontiers in sports and active living}}
  \bibinfo{volume}{2} (\bibinfo{year}{2020}), \bibinfo{pages}{50}.
\newblock


\bibitem[{OptiTrack}(2019)]%
        {skeletontracking}
\bibfield{author}{\bibinfo{person}{{OptiTrack}}.}
  \bibinfo{year}{2019}\natexlab{}.
\newblock \bibinfo{booktitle}{\emph{Skeleton Tracking}}.
\newblock
\urldef\tempurl%
\url{https://docs.optitrack.com/motive/skeleton-tracking}
\showURL{%
\tempurl}


\bibitem[Ostrek et~al\mbox{.}(2019)]%
        {ostrek2019existing}
\bibfield{author}{\bibinfo{person}{Mirela Ostrek}, \bibinfo{person}{Helge
  Rhodin}, \bibinfo{person}{Pascal Fua}, \bibinfo{person}{Erich M{\"u}ller},
  {and} \bibinfo{person}{J{\"o}rg Sp{\"o}rri}.}
  \bibinfo{year}{2019}\natexlab{}.
\newblock \showarticletitle{Are existing monocular computer vision-based 3D
  Motion capture approaches ready for deployment? A methodological study on the
  example of alpine skiing}.
\newblock \bibinfo{journal}{\emph{Sensors}} \bibinfo{volume}{19},
  \bibinfo{number}{19} (\bibinfo{year}{2019}), \bibinfo{pages}{4323}.
\newblock


\bibitem[Pavlakos et~al\mbox{.}(2019)]%
        {pavlakos2019expressive}
\bibfield{author}{\bibinfo{person}{Georgios Pavlakos},
  \bibinfo{person}{Vasileios Choutas}, \bibinfo{person}{Nima Ghorbani},
  \bibinfo{person}{Timo Bolkart}, \bibinfo{person}{Ahmed~AA Osman},
  \bibinfo{person}{Dimitrios Tzionas}, {and} \bibinfo{person}{Michael~J
  Black}.} \bibinfo{year}{2019}\natexlab{}.
\newblock \showarticletitle{Expressive body capture: 3d hands, face, and body
  from a single image}. In \bibinfo{booktitle}{\emph{Proceedings of the
  IEEE/CVF conference on computer vision and pattern recognition}}.
  \bibinfo{pages}{10975--10985}.
\newblock


\bibitem[Pavllo et~al\mbox{.}(2019)]%
        {pavllo20193d}
\bibfield{author}{\bibinfo{person}{Dario Pavllo}, \bibinfo{person}{Christoph
  Feichtenhofer}, \bibinfo{person}{David Grangier}, {and}
  \bibinfo{person}{Michael Auli}.} \bibinfo{year}{2019}\natexlab{}.
\newblock \showarticletitle{3d human pose estimation in video with temporal
  convolutions and semi-supervised training}. In
  \bibinfo{booktitle}{\emph{Proceedings of the IEEE/CVF conference on computer
  vision and pattern recognition}}. \bibinfo{pages}{7753--7762}.
\newblock


\bibitem[Qiu et~al\mbox{.}(2019)]%
        {qiu2019cross}
\bibfield{author}{\bibinfo{person}{Haibo Qiu}, \bibinfo{person}{Chunyu Wang},
  \bibinfo{person}{Jingdong Wang}, \bibinfo{person}{Naiyan Wang}, {and}
  \bibinfo{person}{Wenjun Zeng}.} \bibinfo{year}{2019}\natexlab{}.
\newblock \showarticletitle{Cross view fusion for 3d human pose estimation}. In
  \bibinfo{booktitle}{\emph{Proceedings of the IEEE/CVF international
  conference on computer vision}}. \bibinfo{pages}{4342--4351}.
\newblock


\bibitem[Radford et~al\mbox{.}(2021)]%
        {radford2021learning}
\bibfield{author}{\bibinfo{person}{Alec Radford}, \bibinfo{person}{Jong~Wook
  Kim}, \bibinfo{person}{Chris Hallacy}, \bibinfo{person}{Aditya Ramesh},
  \bibinfo{person}{Gabriel Goh}, \bibinfo{person}{Sandhini Agarwal},
  \bibinfo{person}{Girish Sastry}, \bibinfo{person}{Amanda Askell},
  \bibinfo{person}{Pamela Mishkin}, \bibinfo{person}{Jack Clark},
  {et~al\mbox{.}}} \bibinfo{year}{2021}\natexlab{}.
\newblock \showarticletitle{Learning transferable visual models from natural
  language supervision}. In \bibinfo{booktitle}{\emph{International conference
  on machine learning}}. PMLR, \bibinfo{pages}{8748--8763}.
\newblock


\bibitem[Raskar et~al\mbox{.}(2007)]%
        {raskar2007prakash}
\bibfield{author}{\bibinfo{person}{Ramesh Raskar}, \bibinfo{person}{Hideaki
  Nii}, \bibinfo{person}{Bert Dedecker}, \bibinfo{person}{Yuki Hashimoto},
  \bibinfo{person}{Jay Summet}, \bibinfo{person}{Dylan Moore},
  \bibinfo{person}{Yong Zhao}, \bibinfo{person}{Jonathan Westhues},
  \bibinfo{person}{Paul Dietz}, \bibinfo{person}{John Barnwell},
  {et~al\mbox{.}}} \bibinfo{year}{2007}\natexlab{}.
\newblock \showarticletitle{Prakash: lighting aware motion capture using
  photosensing markers and multiplexed illuminators}.
\newblock \bibinfo{journal}{\emph{ACM Transactions on Graphics (TOG)}}
  \bibinfo{volume}{26}, \bibinfo{number}{3} (\bibinfo{year}{2007}),
  \bibinfo{pages}{36--es}.
\newblock


\bibitem[Ray et~al\mbox{.}(2023)]%
        {ray2023pressim}
\bibfield{author}{\bibinfo{person}{Lala Shakti~Swarup Ray}, \bibinfo{person}{Bo
  Zhou}, \bibinfo{person}{Sungho Suh}, {and} \bibinfo{person}{Paul Lukowicz}.}
  \bibinfo{year}{2023}\natexlab{}.
\newblock \showarticletitle{PresSim: An End-to-end Framework for Dynamic Ground
  Pressure Profile Generation from Monocular Videos Using Physics-based 3D
  Simulation}. In \bibinfo{booktitle}{\emph{2023 IEEE International Conference
  on Pervasive Computing and Communications Workshops and other Affiliated
  Events (PerCom Workshops)}}. IEEE, \bibinfo{pages}{484--489}.
\newblock


\bibitem[Reilink et~al\mbox{.}(2013)]%
        {reilink20133d}
\bibfield{author}{\bibinfo{person}{Rob Reilink}, \bibinfo{person}{Stefano
  Stramigioli}, {and} \bibinfo{person}{Sarthak Misra}.}
  \bibinfo{year}{2013}\natexlab{}.
\newblock \showarticletitle{3D position estimation of flexible instruments:
  marker-less and marker-based methods}.
\newblock \bibinfo{journal}{\emph{International journal of computer assisted
  radiology and surgery}}  \bibinfo{volume}{8} (\bibinfo{year}{2013}),
  \bibinfo{pages}{407--417}.
\newblock


\bibitem[Sigal(2021)]%
        {sigal2021human}
\bibfield{author}{\bibinfo{person}{Leonid Sigal}.}
  \bibinfo{year}{2021}\natexlab{}.
\newblock \showarticletitle{Human pose estimation}.
\newblock In \bibinfo{booktitle}{\emph{Computer Vision: A Reference Guide}}.
  \bibinfo{publisher}{Springer}, \bibinfo{pages}{573--592}.
\newblock


\bibitem[Sigal et~al\mbox{.}(2010)]%
        {sigal2010humaneva}
\bibfield{author}{\bibinfo{person}{Leonid Sigal}, \bibinfo{person}{Alexandru~O
  Balan}, {and} \bibinfo{person}{Michael~J Black}.}
  \bibinfo{year}{2010}\natexlab{}.
\newblock \showarticletitle{Humaneva: Synchronized video and motion capture
  dataset and baseline algorithm for evaluation of articulated human motion}.
\newblock \bibinfo{journal}{\emph{International journal of computer vision}}
  \bibinfo{volume}{87}, \bibinfo{number}{1-2} (\bibinfo{year}{2010}),
  \bibinfo{pages}{4}.
\newblock


\bibitem[Trumble et~al\mbox{.}(2017)]%
        {trumble2017total}
\bibfield{author}{\bibinfo{person}{Matthew Trumble}, \bibinfo{person}{Andrew
  Gilbert}, \bibinfo{person}{Charles Malleson}, \bibinfo{person}{Adrian
  Hilton}, {and} \bibinfo{person}{John Collomosse}.}
  \bibinfo{year}{2017}\natexlab{}.
\newblock \showarticletitle{Total capture: 3d human pose estimation fusing
  video and inertial sensors}. In \bibinfo{booktitle}{\emph{Proceedings of 28th
  British Machine Vision Conference}}. \bibinfo{pages}{1--13}.
\newblock


\bibitem[{University of Cyprus}(2023)]%
        {dance_db}
\bibfield{author}{\bibinfo{person}{{University of Cyprus}}.}
  \bibinfo{year}{2023}\natexlab{}.
\newblock \bibinfo{title}{Dance Motion Capture Database}.
\newblock \bibinfo{howpublished}{\url{http://dancedb.cs.ucy.ac.cy}}.
\newblock
\newblock
\shownote{Motion capture data used in this work were obtained from the Dance
  Motion Capture Database of the University of Cyprus}.


\bibitem[Van~der Kruk and Reijne(2018)]%
        {van2018accuracy}
\bibfield{author}{\bibinfo{person}{Eline Van~der Kruk} {and}
  \bibinfo{person}{Marco~M Reijne}.} \bibinfo{year}{2018}\natexlab{}.
\newblock \showarticletitle{Accuracy of human motion capture systems for sport
  applications; state-of-the-art review}.
\newblock \bibinfo{journal}{\emph{European journal of sport science}}
  \bibinfo{volume}{18}, \bibinfo{number}{6} (\bibinfo{year}{2018}),
  \bibinfo{pages}{806--819}.
\newblock


\bibitem[Wang et~al\mbox{.}(2021)]%
        {wang2021deep}
\bibfield{author}{\bibinfo{person}{Jinbao Wang}, \bibinfo{person}{Shujie Tan},
  \bibinfo{person}{Xiantong Zhen}, \bibinfo{person}{Shuo Xu},
  \bibinfo{person}{Feng Zheng}, \bibinfo{person}{Zhenyu He}, {and}
  \bibinfo{person}{Ling Shao}.} \bibinfo{year}{2021}\natexlab{}.
\newblock \showarticletitle{Deep 3D human pose estimation: A review}.
\newblock \bibinfo{journal}{\emph{Computer Vision and Image Understanding}}
  \bibinfo{volume}{210} (\bibinfo{year}{2021}), \bibinfo{pages}{103225}.
\newblock


\bibitem[Wu et~al\mbox{.}(2019)]%
        {wu2019detectron2}
\bibfield{author}{\bibinfo{person}{Yuxin Wu}, \bibinfo{person}{Alexander
  Kirillov}, \bibinfo{person}{Francisco Massa}, \bibinfo{person}{Wan-Yen Lo},
  {and} \bibinfo{person}{Ross Girshick}.} \bibinfo{year}{2019}\natexlab{}.
\newblock \bibinfo{title}{Detectron2}.
\newblock
  \bibinfo{howpublished}{\url{https://github.com/facebookresearch/detectron2}}.
\newblock


\bibitem[Yi et~al\mbox{.}(2022a)]%
        {yi2022self}
\bibfield{author}{\bibinfo{person}{Chunzhi Yi}, \bibinfo{person}{Baichun Wei},
  \bibinfo{person}{Zhen Ding}, \bibinfo{person}{Chifu Yang},
  \bibinfo{person}{Zhiyuan Chen}, {and} \bibinfo{person}{Feng Jiang}.}
  \bibinfo{year}{2022}\natexlab{a}.
\newblock \showarticletitle{A self-aligned method of IMU-based 3-DoF lower-limb
  joint angle estimation}.
\newblock \bibinfo{journal}{\emph{IEEE Transactions on Instrumentation and
  Measurement}}  \bibinfo{volume}{71} (\bibinfo{year}{2022}),
  \bibinfo{pages}{1--10}.
\newblock


\bibitem[Yi et~al\mbox{.}(2022b)]%
        {yi2022physical}
\bibfield{author}{\bibinfo{person}{Xinyu Yi}, \bibinfo{person}{Yuxiao Zhou},
  \bibinfo{person}{Marc Habermann}, \bibinfo{person}{Soshi Shimada},
  \bibinfo{person}{Vladislav Golyanik}, \bibinfo{person}{Christian Theobalt},
  {and} \bibinfo{person}{Feng Xu}.} \bibinfo{year}{2022}\natexlab{b}.
\newblock \showarticletitle{Physical inertial poser (pip): Physics-aware
  real-time human motion tracking from sparse inertial sensors}. In
  \bibinfo{booktitle}{\emph{Proceedings of the IEEE/CVF Conference on Computer
  Vision and Pattern Recognition}}. \bibinfo{pages}{13167--13178}.
\newblock


\bibitem[Zhou et~al\mbox{.}(2023)]%
        {zhou2023mocapose}
\bibfield{author}{\bibinfo{person}{Bo Zhou}, \bibinfo{person}{Daniel Geissler},
  \bibinfo{person}{Marc Faulhaber}, \bibinfo{person}{Clara~Elisabeth Gleiss},
  \bibinfo{person}{Esther~Friederike Zahn}, \bibinfo{person}{Lala Shakti~Swarup
  Ray}, \bibinfo{person}{David Gamarra}, \bibinfo{person}{Vitor~Fortes Rey},
  \bibinfo{person}{Sungho Suh}, \bibinfo{person}{Sizhen Bian}, {et~al\mbox{.}}}
  \bibinfo{year}{2023}\natexlab{}.
\newblock \showarticletitle{MoCaPose: Motion Capturing with Textile-integrated
  Capacitive Sensors in Loose-fitting Smart Garments}.
\newblock \bibinfo{journal}{\emph{Proceedings of the ACM on Interactive,
  Mobile, Wearable and Ubiquitous Technologies}} \bibinfo{volume}{7},
  \bibinfo{number}{1} (\bibinfo{year}{2023}), \bibinfo{pages}{1--40}.
\newblock


\end{thebibliography}

% \appendix
\end{document}